\documentclass{article}
\usepackage{amsfonts,pstricks}
\usepackage{graphicx}
\newcommand{\R}{\mathbb{R}}

\begin{document}

\title{Ward's Hierarchical Clustering Method: Clustering Criterion and 
Agglomerative Algorithm}

\author{Fionn Murtagh (1) and Pierre Legendre (2) \\
(1) Science Foundation Ireland, Wilton Park House, \\
Wilton Place, Dublin 2, Ireland (fmurtagh@acm.org) \\
(2) D\'epartement de sciences biologiques, Universit\'e 
de Montr\'eal,\\ 
C.P. 6128 succursale Centre-ville, Montr\'eal, Qu\'ebec, \\
Canada H3C 3J7 (pierre.legendre@umontreal.ca)}

\maketitle

\begin{abstract}
The Ward error sum of squares hierarchical clustering method 
has been very widely used since its first description by Ward
in a 1963 publication.  It has also been generalized in various 
ways.  However there are different interpretations in the 
literature and there are different implementations of the Ward
agglomerative algorithm in commonly used software systems, including 
differing expressions of the agglomerative criterion.
Our survey work and case studies will be useful for all those involved in 
developing software for 
data analysis using Ward's hierarchical clustering method.
\end{abstract}

\noindent
{\bf Keywords:}
Hierarchical clustering, Ward, Lance-Williams, minimum variance.


\section{Introduction}

In the literature and in software packages there is confusion in 
regard to what is termed the Ward hierarchical clustering method.  
This relates to any and possibly all of the following: (i) input
dissimilarities, whether squared or not; (ii) output dendrogram 
heights and whether or not their square root is used; and (iii) 
there is a subtle but important difference that we have found 
in the loop structure of the stepwise dissimilarity-based 
agglomerative algorithm.   Our main objective in this work is 
to warn users of hierarchical clustering about this, to raise
awareness about these distinctions or differences, and to 
urge users to check what their favorite software package is doing. 

In R, the function {\tt hclust} of {\tt stats} with the 
{\tt method="ward"} 
option produces results that correspond to a Ward method 
(Ward\footnote{This article is dedicated to Joe H.\ Ward Jr., 
who died on 23 June 2011, aged 84.}, 1963)
described in terms 
of a Lance-Williams updating formula using a sum of dissimilarities,
which produces updated dissimilarities.  This is the implementation
used by, for example, Wishart (1969), Murtagh (1985) on whose code the 
{\tt hclust} implementation is based, Jain and Dubes (1988), Jambu (1989), 
in XploRe (2007), in Clustan (www.clustan.com), and elsewhere.

An important issue though is the form of input that is necessary to 
give Ward's method.   For an input data matrix, {\tt x}, in R's
{\tt hclust} function the following command is required:
\verb+hclust(dist(x)^2,method="ward")+.  In later sections 
(in particular, section \ref{sect31}) of this 
article we explain just why the squaring of the distances is 
 a requirement for the Ward method.  In section \ref{sect4} (Experiment
4) it is discussed why we may wish to take the square roots of the 
agglomeration, or dendrogram node height, values.   

In R, the {\tt agnes} function of {\tt cluster} with the 
{\tt method="ward"} option is also 
presented as the Ward method in Kaufman and 
Rousseeuw (1990), Legendre and Legendre (2012), among others.   
A formally similar algorithm is used, based on the Lance and 
Williams (1967) recurrence.  

Lance and Williams (1967) did not themselves consider the
Ward method, which instead was first investigated by Wishart (1969).

What is at issue for us here starts with how {\tt hclust} and 
{\tt agnes} give different outputs when applied to the same 
dissimilarity matrix as input.  What therefore explains
the formal similarity in terms of criterion and algorithms, yet 
at the same time yields outputs that are different?

\section{Ward's Agglomerative Hierarchical Clustering Method}

\subsection{Some Definitions}
\label{defns}

We recall that a distance is a positive, definite, symmetric mapping 
of a pair of observation vectors onto the positive reals which in addition
satisfies the triangular inequality.   For observations $i$, $j$, $k$ we have:
$d(i,j) > 0; d(i,j) = 0 \Longleftrightarrow i = j; d(i,j) = d(j,i); 
d(i,j) \leq d(i,k) + d(k,j)$.  For observation set, $I$, with $i, j, k 
\in I$ we can write the distance as a mapping from the Cartesian product
of the observation set into the positive reals:  
$ d : I \times I \longrightarrow \R^+$.   

A dissimilarity is usually taken as a distance but without the triangular
inequality ($d(i,j) \leq d(i,k) + d(k,j), \forall i, j, k$).    
Lance and Williams, 1967, use the term ``an $(i,j)$-measure'' for a
dissimilarity.

An ultrametric, or tree distance, which defines a hierarchical clustering
(and also an ultrametric topology, which goes beyond a metric geometry, 
or a p-adic number system) differs from a distance in that the strong
triangular inequality is instead satisfied.  This inequality, also commonly 
called the ultrametric inequality, is: $d(i,j) \leq \max \{ d(i,k), d(k,j) \}$.

For observations $i$ in a cluster $q$, and a distance $d$ (which can 
potentially be relaxed to a dissimilarity) we have the following definitions. 
We may want to consider a mass or weight associated with observation $i$, 
$p(i)$.  Typically we take $p(i) = 1/|q|$ when $i \in q$, i.e.\ 1 over 
cluster cardinality of the relevant cluster.

With the context being clear, let $q$ denote the cluster (a set) 
as well as the cluster's center.  We have this center defined as 
$q = 1/|q| \sum_{i \in q} i$.   Furthermore, and again the context makes this
clear, we have $i$ used for the observation label, or index, among 
all observations, and the observation vector.   

Some further definitions follow. 

\begin{description}
\item[ ] Error sum of squares: $\sum_{i \in q} d^2(i,q)$.  

\item[ ] Variance (or centered sum of squares): 
$1/|q| \sum_{i \in q} d^2(i,q)$.
  
\item[ ] Inertia: $\sum_{i \in q} p(i) d^2(i,q)$
which becomes variance if $p(i) = 1/|q|$, and becomes error sum of 
squares if $p(i) = 1$.  

\item[ ] Euclidean distance squared 
using norm $\| . \|$: if $i, i' \in \R^{|J|}$, i.e.\ these observations
have values on attributes $j \in \{1, 2, \dots , |J| \}$, $J$ is the 
attribute set, $| . |$ denotes 
cardinality, then $d^2(i,i') = \| i - i' \|^2 = \sum_j (i_j - i'_j)^2$.   
\end{description} 

Consider now a set of masses, or weights, $m_i$ for observations $i$.  
Following 
Benz\'ecri (1976, p.\ 185), the centered moment of order 2, $M^2(I)$ of 
the cloud (or set) of observations $i, i \in I$, is written: 
$M^2(I) = \sum_{i \in I} m_i \| i - g \|^2$ where the center of gravity of
the system is $g = \sum_i m_i i / \sum_i m_i$ .  The variance, $V^2(I)$, 
is $V^2(I) = M^2(I)/m_I$, where $m_I$ is the total mass of the cloud.   
Due to Huyghen's theorem the following can be 
shown (Benz\'ecri, 1976, p.\ 186) for clusters $q$ whose union make up 
the partition, $Q$: 

$$ M^2(Q) = \sum_{q \in Q} m_q \| q - g \|^2 $$
$$ M^2(I) = M^2(Q) + \sum_{q \in Q} M^2(q) $$
$$ V(Q) = \sum_{q \in Q} \frac{m_q}{m_I} \| q - g \|^2 $$
$$ V(I) = V(Q) + \sum_{q \in Q} \frac{m_q}{m_I} V(q) $$

The $V(Q)$ and $V(I)$ definitions here are discussed in Jambu (1978,
pp.\ 154--155).   The last of the above can be seen to decompose 
(additively) total variance of the cloud $I$ into (first term on the 
right hand side) variance of the cloud of cluster centers ($q \in Q$), 
and summed variances of the clusters.  We can consider this last of 
the above relations as: $T = B + W$, where $B$ is the between clusters
variance, and $W$ is the summed within clusters variance.  

A range of variants of the agglomerative clustering criterion and 
algorithm are discussed by Jambu (1978).  These 
include: minimum of the centered order 2 moment of the union of two 
clusters (p.\ 156); minimum variance of the union of two clusters 
(p.\ 156); maximum of the centered order 2 moment of a partition (p.\
157);  and maximum of the centered order 2 moment of a partition (p.\
158).  Jambu notes that these criteria for maximization of the centered
order 2 moment, or variance, of a partition, were developed and used 
by numerous authors, with some of these authors introducing modifications
(such as the use of particular metrics).   Among authors referred to 
are Ward (1963), Orlocci, Wishart (1969), and various others. 

\subsection{Alternative Expressions for the Variance}

In the previous section, the variance was written as
$1/|q| \sum_{i \in q} d^2(i,q)$.
This is the so-called population variance.  
When viewed in statistical terms, where an unbiased estimator 
of the variance is needed, we require the sample variance:
$1/(|q|-1) \sum_{i \in q} d^2(i,q)$.  
The population quantity is used in Murtagh (1985).  The sample 
statistic is used in Le Roux and Rouanet (2004), and by 
Legendre and Legendre (2012).  

The sum of squares, $\sum_{i \in q} d^2(i,q)$, can be written
in terms of all pairwise distances: 

$\sum_{i \in q} d^2(i,q) = 1/|q| \sum_{i, i' \in q, i < i'} d^2(i,i')$.
This is proved as follows (see, e.g., Legendre and Fortin, 
2010).  Write 

$\frac{1}{|q|} \sum_{i, i' \in q, i < i'} d^2(i,i') = \frac{1}{|q|} 
\sum_{i, i' \in q, i < i'} (i - i')^2$

$= \frac{1}{|q|} \sum_{i, i' \in q, i < i'} (i - q - (i' - q) )^2$ where, as 
already noted in section \ref{defns}, $ q = \frac{1}{|q|}  \sum_{i \in q} i$.

$= \frac{1}{|q|} \sum_{i, i' \in q, i < i'} \left( (i - q)^2 + (i' - q)^2 - 2 
   (i - q) (i' - q) \right)$

$= \frac{1}{2} \frac{1}{|q|} 
\sum_{i \in q} \sum_{i' \in q} \left( (i - q)^2 + (i' - q)^2 - 2 (i - q) (i' - q) \right)$

$= \frac{1}{2} \frac{1}{|q|} \left( 2 \sum_{i \in q}  (i - q)^2 \right) 
- \frac{1}{2} \frac{1}{|q|} \left( \sum_{i \in q} \sum_{i' \in q} 2 (i - q) (i' - q) \right)$ 

By writing out the right hand term, we see that it equals 0.   Hence our result.  

As noted in Legendre and Legendre (2012) there are many other 
alternative expressions for 
calculating $\sum_{i \in q} d^2(i,q)$, such as using the trace of a particular 
distance matrix, and the sum of eigenvalues of a principal coordinate analysis of the 
distance matrix.   The latter is invoking what is known as the Parseval relation, i.e.\
the equivalence of the norms of vectors in inner product spaces that can be orthonormally 
transformed, one space to the other.  

\subsection{Lance-Williams Dissimilarity Update Formula}
\label{sect22}

Lance and Williams (1967) established a succinct
form for the update of dissimilarities following an agglomeration.  
The parameters used in the update formula are dependent on the cluster
criterion value.  Consider clusters (including possibly singletons)
$i$ and $j$ being agglomerated to form cluster $ i \cup j$, and then consider 
the re-defining of dissimilarity relative to an external cluster
(including again possibly a singleton), $k$.  We have: 

$$d(i \cup j, k) = a(i) \cdot d(i,k) + a(j) \cdot d(j,k) + b \cdot d(i,j) 
+ c \cdot | d(i,k) - d(j,k) | $$
\noindent
where $d$ is the dissimilarity used -- which does not have to be a 
Euclidean distance to start with, insofar as the Lance and Williams 
formula can be used as a repeatedly executed recurrence, without 
reference to any other or separate criterion; 
coefficients $a(i), a(j), b, c$ are 
defined with reference to the clustering criterion used (see tables of 
these coefficients in Murtagh, 1985, p.\ 68; Jambu, 1989, p.\ 366); and 
$| . |$ denotes absolute value.   

The Lance-Williams recurrence formula considers dissimilarities and not 
dissimilarities squared.  

The original Lance and Williams (1967) paper does not consider the Ward 
criterion.  
It does however note that it allows one to 
``generate an infinite set of new strategies'' for agglomerative hierarchical
clustering.  Wishart (1969) brought the Ward criterion into the Lance-Williams
algorithmic framework.  

Even starting the agglomerative 
process 
with a Euclidean distance will not avoid the fact that the inter-cluster 
(non-singleton, i.e.\ with 2 or more members) dissimilarity does not 
respect the triangular inequality, and hence it does not respect this  
Euclidean metric property.  

\subsection{Generalizing Lance-Williams}


The Lance and Williams recurrence formula has been generalized in various ways.
See e.g.\ Batagelj (1988) who discusses what he terms ``generalized 
Ward clustering'' which includes agglomerative criteria based on 
variance, inertia and weighted increase in variance.  

Jambu (1989, pp.\
356 et seq.) considers the following cluster criteria and associated 
Lance-Williams update formula in the generalized 
Ward framework: centered order 2 moment of a partition; 
variance of a partition; centered order 2 moment of the union of two classes;
and variance of the union of two classes.

If using a Euclidean distance, the Murtagh (1985) and the Jambu (1989) 
Lance-Williams update formulas for variance and related criteria (as 
discussed by Jambu, 1989) are associated with an alternative agglomerative 
hierarchical clustering algorithm which defines cluster centers following
each agglomeration, and 
thus does not require use of the Lance-Williams update formula.  
The same is true for hierarchical agglomerative clustering based on 
median and centroid criteria.  

As noted, the Lance-Williams update formula uses a dissimilarity, $d$.  
Sz\'ekely and Rizzo (2005) consider higher order powers of this, in the 
Ward context:  
``Our proposed method extends Ward's minimum variance method.  
Ward's method minimizes the increase in total within-cluster sum 
of squared error.  This increase is proportional to the squared 
Euclidean distance between cluster centers.   In contrast to Ward's 
method, our cluster distance is based on Euclidean distance, rather 
than squared Euclidean distance.  More generally, we define ... an 
objective function and cluster distance in terms of any power $\alpha$ 
of Euclidean distance in the interval (0,2] ... Ward's mimimum variance 
method is obtained as the special case when $\alpha = 2$.''

Then it is indicated what beneficial properties the case of $\alpha = 1$ 
has, including: Lance-Williams form, ultrametricity and reducibility, 
space-dilation, and computational tractability.    
In Sz\'ekely and Rizzo (2005, p. 164) it is stated that 
``We have shown that'' the $\alpha = 1$ case, 
rather than $\alpha = 2$, gives ``a method that applies to a more general 
class of clustering problems'', and this finding is further emphasized in 
their conclusion. 
Notwithstanding this finding of Sz\'ekely and Rizzo (2005), viz.\
that the $\alpha = 1$ case is best, in this work our interest 
remains with the $\alpha = 2$ Ward method.  

Our objective in this section has been to discuss some of the ways
that the Ward method has been generalized.   We now, in the next section,
come to our central theme in this article.  

\section{Implementations of Ward's Method}
\label{sect3}

We now come to the central part of our work, distinguishing in 
 subsections \ref{sect31} and \ref{sect32} how we can arrive at subtle but 
important differences in relation to how the Ward method, or what is 
said to be the Ward method, is understood in practice, and put into 
software code.  We consider: 
 data inputs, the main loop structure of the 
agglomerative, dissimilarity-based algorithms, and the output dendrogram 
node heights.   The subtle but important differences that we 
uncover are further explored and exemplified in section \ref{sect4}.  

Consider hierarchical clustering in the following form.  On an 
observation set, $I$, define a dissimilarity measure.  Set each of 
the observations, $i, j, k,$ etc.\ $\in I$ to be a singleton cluster.
Agglomerate the closest (i.e. least dissimilarity) pair of clusters, 
deleting the agglomerands, or agglomerated clusters.  Redefine
the inter-cluster dissimilarities with respect to the newly created 
cluster.  If $n$ is the cardinality of observation set $I$ then this 
agglomerative hierarchical clustering algorithm completes in $n- 1$
agglomerative steps.   


Through use of the Lance-Williams update formula, we will focus on
the updated dissimilarities relative to a newly created cluster.  
Unexpectedly in this work, we found a need to focus also on the 
form of input dissimilarities.  

\subsection{The Minimand or Cluster Criterion Optimized}
\label{optim}


The function to be minimized, or minimand, 
in the Ward2 case (see subsection \ref{sect32}), 
as stated by Kaufman and Rousseeuw
(1990, p.\ 230, cf.\  relation (22)) is: 

\begin{equation}
D(c_1, c_2) = \delta^2(c_1, c_2) = \frac{ |c_1| |c_2|}{|c_1|+ |c_2|} 
\| c_1 - c_2\|^2
\label{wardcrit2}
\end{equation}

\noindent 
whereas for the Ward1 case, as discussed in subsection 
\ref{sect31}, we have: 

\begin{equation}
\delta(c_1, c_2) = \frac{ |c_1| |c_2|}{|c_1|+ |c_2|} 
\| c_1 - c_2\|^2
\label{wardcrit1}
\end{equation}

It is clear therefore that the same criterion is being optimized. Both 
implementations minimize the change in variance, or the error sum of squares.

Since the 
error sum of squares, or minimum variance, or other related, criteria 
are not optimized precisely in practice, due to being NP-complete
optimization problems (implying therefore that only exponential search 
in the solution space will guarantee an optimal solution), we are content 
with good heuristics in practice, i.e.\ sub-optimal solutions.  Such a
heuristic is the sequence of two-way agglomerations carried out by a hierarchical
clustering algorithm.  

In either form the criterion (\ref{wardcrit2}), (\ref{wardcrit1}) is characterized 
in Le Roux and Rouanet (2004, p.\ 109) as the variance index; the inertia index; the centered 
moment of order 2; the Ward index (citing Ward, 1963); and the following --
given two classes $c_1$ and $c_2$, the variance index is the contribution 
of the dipole of the class centers, denoted as in (\ref{wardcrit1}).  The 
resulting clustering is termed Euclidean classification by Le Roux and 
Rouanet (2004).  

As noted by Le Roux and Rouanet (2004, p.\ 110), the variance index (as 
they term it) (\ref{wardcrit1}) does not itself satisfy the triangular 
inequality.  To see this just take equidistant clusters with $|c_1| = 3, 
|c_2| = 2, |c_3| = 1$.   


\subsection{Implementation of Ward: Ward1}
\label{sect31}

We start with (let us term it) the Ward1
algorithm as given in Murtagh (1985).  

It was initially Wishart (1969) who wrote the Ward algorithm in 
terms of the Lance-Williams update formula. In Wishart
(1969) the Lance-Williams formula is written in terms of squared
dissimilarities, in a way that is formally identical 
to the following.  

\noindent
Cluster update formula:

\begin{displaymath}
 \delta(i \cup i', i'') = 
\end{displaymath}
\begin{displaymath}
\frac{w_i + w_i''} {w_i + w_{i'} + w_{i''}} \delta(i, i'') +
\frac{w_i' + w_i''}{w_i + w_{i'} + w_{i''}} \delta(i', i'') -
\frac{w_i''}{w_i + w_{i'} + w_{i''}} \delta(i, i') 
\end{displaymath}
\begin{equation}
\mbox{ and } w_{i \cup i'} = w_i + w_{i'}  
\label{ward1}
\end{equation} 

For the optimand of section \ref{optim}, the input dissimilarities 
need to be as follows: 
$\delta(i,i') = \sum_j (x_{ij} - x_{i'j})^2 $.
Note the presence of the {\em squared Euclidean distance} 
in this initial dissimilarity specification.  This is the Ward 
algorithm of Murtagh (1983, 1985 and 2000), and the way that 
\verb+hclust+ in R needs to be used.  
When, however, the
Ward1 algorithm is used with Euclidean distances as the initial
dissimilarities, then the clustering topology can be very different, as will
be seen in Section \ref{sect4}.

The weight $w_i$ is the cluster cardinality, and thus for a 
singleton, $w_i = 1$.  An immediate generalization is to consider 
probabilities given by $w_i = 1/n$.  Generalization to arbitrary 
weights can also be considered. 
Ward implementations that take observation 
weights into account are available in Murtagh (2000).  



$\frac{1}{2} \sum_j (x_{ij} - x_{i'j})^2$, i.e.\
0.5 times Euclidean distances squared, is the sample variance 
(cf.\ section \ref{defns}) of 
the new cluster, $i \cup i'$.  To see this, note that the variance of the new 
cluster $ c$ formed by merging $c_1$ and $c_2$ is 
$( |c_1| \| c_1 - c \|^2 + |c_2| \| c_2 - c \|^2 )/(|c_1| + |c_2|)$ 
where $|c_1|$ is both cardinality and the 
mass of cluster $c_1$, and $\| . \|$ is the Euclidean norm.  The new 
cluster's center of gravity, or mean, is $c = \frac{ |c_1| c_1 + |c_2| c_2}
{|c_1| + |c_2|}$.   By using this expression for the new cluster's center
of gravity (or mean) in the expression given for the variance, we see that we 
can write the variance of the new cluster $c$ combining $c_1$ and $c_2$ 
to be $\frac{ |c_1| |c_2|}{|c_1|+ |c_2|} \| c_1 - c_2\|^2$.   So when 
$|c_1| = |c_2|$ we have the stated result, i.e.\ the variance of the new
cluster equaling 0.5 times Euclidean distances squared.

The criterion that is optimized arises from the foregoing discussion 
(previous paragraph), 
i.e.\ the variance of the dipole formed by the agglomerands.  This is 
the variance of new cluster $c$ minus the variances of (now agglomerated) 
clusters $c_1$ and $c_2$, which we can write as Var$(c) - $Var$(c_1) - 
$Var$(c_2)$.   The variance of the partition containing $c$ necessarily 
decreases, so we need to minimize this decrease when carrying out an 
agglomeration.   

Murtagh (1985) also shows how this optimization criterion is viewed
as achieving the (next possible) partition with maximum between-cluster 
variance.  Maximizing between-cluster variance is the same as minimizing 
within-cluster variance, arising out of Huyghen's variance (or inertia) 
decomposition theorem.  
With reference to section \ref{defns} we are minimizing the change in 
$B$, hence maximizing $B$, and hence minimizing $W$.

Jambu (1978, p.\ 157) calls the Ward1 algorithm the maximum centered order
2 moment of a partition (cf.\ section \ref{defns} above).  The 
criterion is denoted by him as $\delta_{\mbox{mot}}$.  

\subsection{Implementation of Ward: Ward2}
\label{sect32}

We now look at the Ward2 algorithm discussed in Kaufman and Rousseeuw
(1990), and Legendre and Legendre (2012).  

At each agglomerative step, the extra sum of squares caused by agglomerating 
clusters is minimized, exactly as we have seen for the Ward1 algorithm 
above.   We have the following. 

\noindent
Cluster update formula:

\begin{displaymath} 
\delta(i \cup i', i'') =
\end{displaymath}
\begin{displaymath}
\big( \frac{w_i + w_i''} {w_i + w_{i'} + w_{i''}} \delta^2(i, i'') +
\frac{w_i' + w_i''}{w_i + w_{i'} + w_{i''}} \delta^2(i', i'') -
\frac{w_i''}{w_i + w_{i'} + w_{i''}} \delta^2(i, i') \big)^{1/2} 
\end{displaymath}
\begin{equation}
\mbox{ and } w_{i \cup i'} = w_i + w_{i'}
\label{ward2}  
\end{equation} 

Exactly as for Ward1, we have input dissimilarities given by the 
{\em squared Euclidean distance}.  Note though the required form for this, 
in the case of equation \ref{ward2}: 
$\delta^2(i,i') = \sum_j (x_{ij} - x_{i'j})^2 $.
It is such squared Euclidean distances 
that interest us, since our motivation arises from the error sum
of squares criterion.   Note, very importantly, that the $\delta$ function 
is not the same in equation \ref{ward1} and in equation \ref{ward2}; 
this $\delta$ function is, respectively, a squared distance and a distance. 

A second point to note is that equation \ref{ward2} relates to, on the 
right hand side, {\em the square root of a weighted sum of squared distances}.
Consider 
how in equation \ref{ward1} the cluster update formula was in terms of 
{\em a weighted sum of distances}.  

A final point about equation \ref{ward2} is that in the cluster update formula 
it is the set of $\delta$ values that we seek.  

Now let us look further at the relationship between equations \ref{ward2} and 
\ref{ward1}, and show their relationship.  
Rewriting the cluster update formula establishes that we have:

\begin{displaymath} 
\delta^2(i \cup i', i'') =
\end{displaymath}
\begin{equation}
\frac{w_i + w_i''} {w_i + w_{i'} + w_{i''}} \delta^2(i, i'') +
\frac{w_i' + w_i''}{w_i + w_{i'} + w_{i''}} \delta^2(i', i'') -
\frac{w_i''}{w_i + w_{i'} + w_{i''}} \delta^2(i, i')
\end{equation}
 
Let us use the notation $D = \delta^2$ because then, with 

\begin{displaymath}
D(i \cup i', i'') =
\end{displaymath}
\begin{equation}
\frac{w_i + w_i''} {w_i + w_{i'} + w_{i''}} D(i, i'') +
\frac{w_i' + w_i''}{w_i + w_{i'} + w_{i''}} D(i', i'') -
\frac{w_i''}{w_i + w_{i'} + w_{i''}} D(i, i')
\end{equation}

\noindent
we see exactly the form of the Lance-Williams cluster update formula
(section \ref{sect22}).   

Although the agglomerative clustering algorithm is not fully 
specified as such  
in Cailliez and Pag\`es (1976), it appears that the Ward2 algorithm 
is the one attributed to Ward (1963).  See their criterion $d_9$  
(Cailliez and Pag\`es, 1976, pp.\ 531, 571).


With the appropriate choice of $\delta$, 
different for Ward1 and for Ward2, what we have here is the 
identity of the algorithms Ward1 and Ward2, although they are implemented 
to a small extent differently.  We show this as follows.  

Take the Ward2 algorithm one step 
further than above, and write the 
input dissimilarities and cluster update formula using $D$.  We have
the following then. 

Input dissimilarities:  $D(i,i') = \delta^2(i,i') = 
\sum_j (x_{ij} - x_{i'j})^2  $.
Cluster update formula:

\begin{displaymath}
 D(i \cup i', i'') =
\end{displaymath}
\begin{displaymath}
\frac{w_i + w_i''} {w_i + w_{i'} + w_{i''}} D(i, i'') +
\frac{w_i' + w_i''}{w_i + w_{i'} + w_{i''}} D(i', i'') -
\frac{w_i''}{w_i + w_{i'} + w_{i''}} D(i, i')  
\end{displaymath}
\begin{equation} 
\mbox{ and } 
w_{i \cup i'} = w_i + w_{i'}  
\label{ward2b}
\end{equation} 

In this form, equation \ref{ward2b}, implementation Ward2 (equation \ref{ward2})
is {\em identical} to implementation Ward1 (equation \ref{ward1}).   
We conclude that we {\em can} have
Ward1 and Ward2 implementations such that the outputs are identical.  

\section{Case Studies: Ward Implementations and Their Relationships}
\label{sect4}

The hierarchical clustering programs used in this set of case studies are:

\begin{itemize}
\item {\tt hclust} in package {\tt stats}, ``The R Stats Package'',  in R.
Based on code by F.\ Murtagh (Murtagh, 1985), 
included in R by Ross Ihaka. 
\item {\tt agnes} in package {\tt cluster}, ``Cluster Analysis Extended Rousseeuw et al.'', 
in  R, by L. Kaufman and P.J. Rousseeuw.  
\item {\tt hclust.PL}, an extended version of {\tt hclust} in R, by P.\ 
Legendre.  In this function, the Ward1 algorithm is implemented by 
{\tt method="ward.D"} and the Ward2 algorithm by
{\tt method="ward.D2"}.
\end{itemize}

We ensure reproducible results by providing all code used and, 
to begin with, by generating an input data set as follows.  

\begin{verbatim}
# Fix the seed of the random number generator in order
# to have reproducible results. 
set.seed(19037561)
# Create the input matrix to be used. 
y <- matrix(runif(20*4),nrow=20,ncol=4)
# Look at overall mean and column standard deviations.
mean(y); sd(y)
0.4920503   # mean
0.2778538 0.3091678 0.2452009 0.2918480  # std. devs. 
\end{verbatim} 

\subsection{Experiment 1: {\tt agnes} and Ward2 Implementation,
{\tt hclust.PL}}

The R code used is shown in the following, with output produced.  
In all of these experiments, we used the dendrogram node 
heights, associated with the agglomeration criterion values, 
in order to quickly show numerical equivalences.  This is then 
followed up with displays of the dendrograms.  

\begin{verbatim}
# EXPERIMENT 1 ----------------------------------
X.hclustPL.wardD2 = hclust.PL(dist(y),method="ward.D2")
X.agnes.wardD2 = agnes(dist(y),method="ward")

sort(X.hclustPL.wardD2$height)
0.1573864 0.2422061 0.2664122 0.2901741 0.3030634 
0.3083869 0.3589344 0.3830281 0.3832023 0.5753823 
0.6840459 0.7258152 0.7469914 0.7647439 0.8042245 
0.8751259 1.2043397 1.5665054 1.8584163

sort(X.agnes.wardD2$height)
0.1573864 0.2422061 0.2664122 0.2901741 0.3030634 
0.3083869 0.3589344 0.3830281 0.3832023 0.5753823 
0.6840459 0.7258152 0.7469914 0.7647439 0.8042245 
0.8751259 1.2043397 1.5665054 1.8584163

\end{verbatim}

This points to: 
{\tt hclust.PL} with the {\tt method="ward.D2"} option
being identical to:
{\tt agnes} with the {\tt method="ward"} option.

\begin{figure}
\begin{center}
\includegraphics[width=16cm]{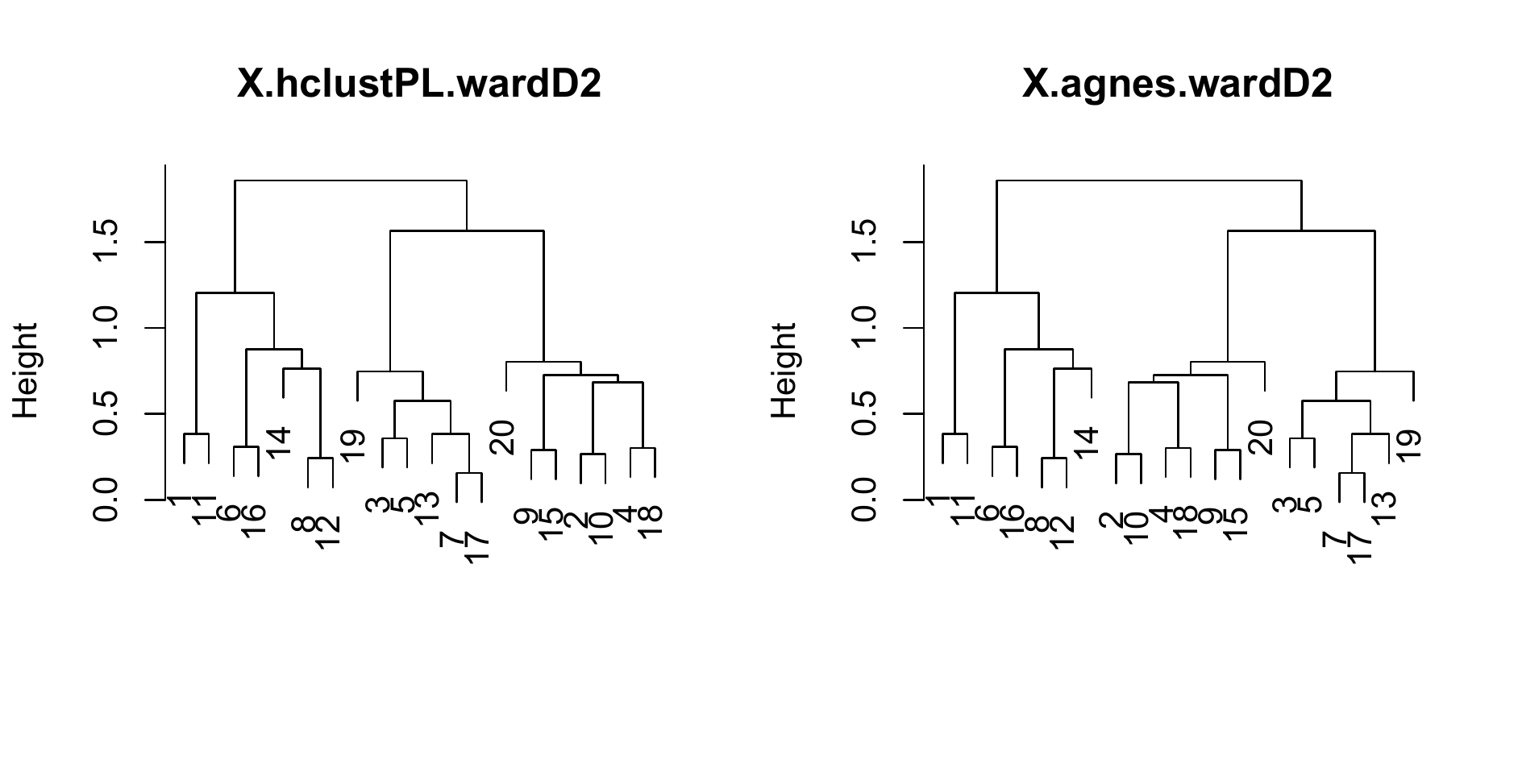}
\end{center}
\caption{Experiment 1 outcome.}
\label{expt1}
\end{figure}

Figure \ref{expt1} displays the outcome, and we see
the same visual result in both cases.  That is, the two 
dendrograms are identical except for inconsequential swiveling 
of nodes.  In group theory terminology we say that the trees
are wreath product invariant.  

To fully complete our reproducibility of research agenda, 
this is the code used to produce Figure \ref{expt1}:

\begin{verbatim}
par(mfrow=c(1,2))
plot(X.hclustPL.wardD2,main="X.hclustPL.wardD2",sub="",xlab="")
plot(X.agnes.wardD2,which.plots=2,main="X.agnes.wardD2",sub="",xlab="")
\end{verbatim} 

\subsection{Experiment 2: {\tt hclust} and Ward1 Implementation,
{\tt hclust.PL}}

Code used is as follows, with output shown.  

\begin{verbatim}

# EXPERIMENT 2 ----------------------------------
X.hclust = hclust(dist(y)^2, method="ward")
X.hclustPL.sq.wardD = hclust.PL(dist(y)^2, method="ward.D")

sort(X.hclust$height)
0.02477046 0.05866380 0.07097546 0.08420102 0.09184743 
0.09510249 0.12883390 0.14671052 0.14684403 0.33106478 
0.46791879 0.52680768 0.55799612 0.58483318 0.64677705 
0.76584542 1.45043423 2.45393902 3.45371103

sort(X.hclustPL.sq.wardD$height)
0.02477046 0.05866380 0.07097546 0.08420102 0.09184743 
0.09510249 0.12883390 0.14671052 0.14684403 0.33106478 
0.46791879 0.52680768 0.55799612 0.58483318 0.64677705 
0.76584542 1.45043423 2.45393902 3.45371103

\end{verbatim}

This points to:
{\tt hclust}, with  {\tt "ward"} option,  on squared input
being identical to: 
{\tt hclust.PL} with {\tt method="ward.D"} option, on squared input.

The clustering levels shown here in Experiment 2 
are the squares of the clustering
levels produced by Experiment 1. 

Figure \ref{expt2} displays the outcome, and we see
the same visual result in both cases.   
This is the code used to produce Figure \ref{expt2}:

\begin{figure}
\begin{center}
\includegraphics[width=16cm]{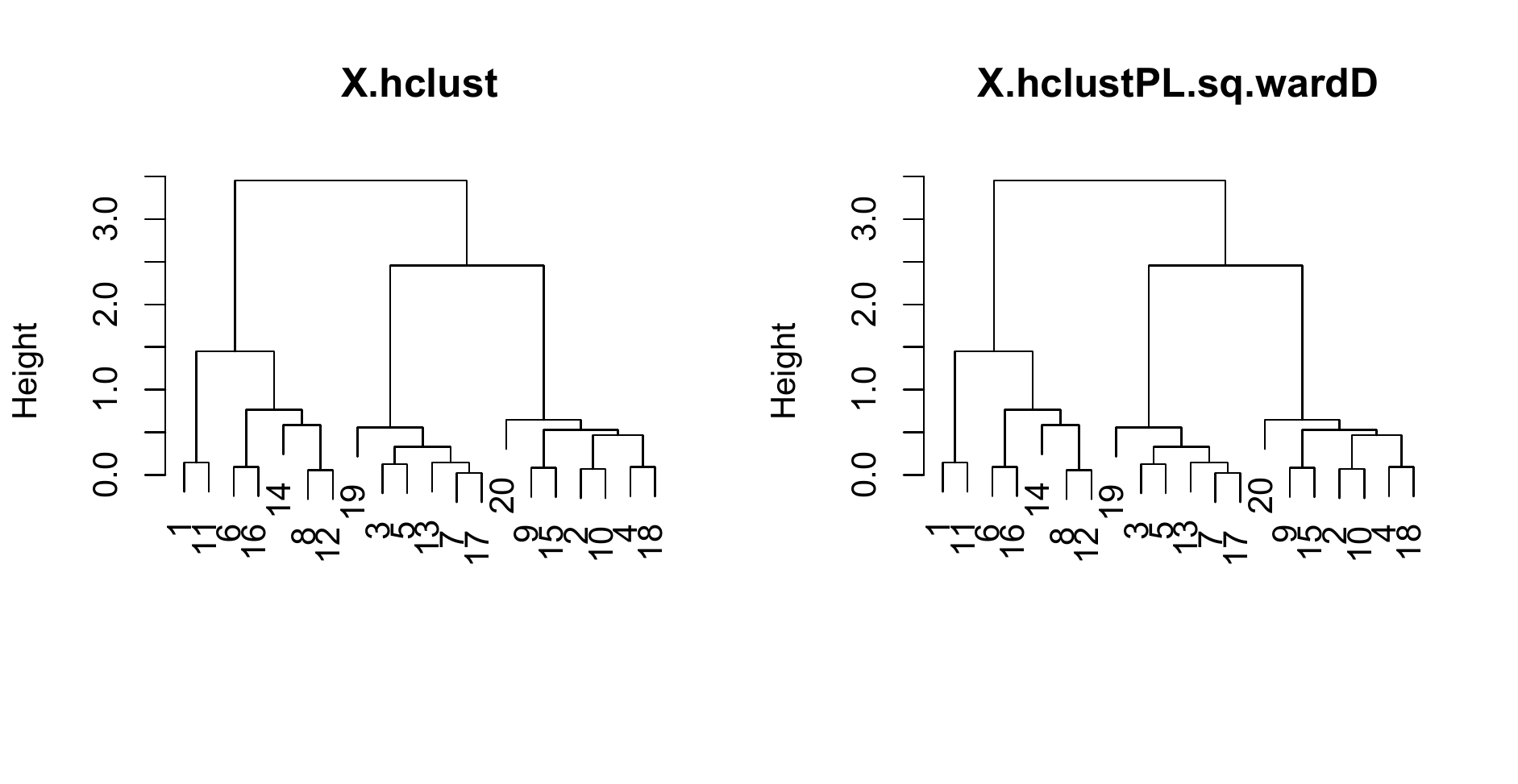}
\end{center}
\caption{Experiment 2 outcome.}
\label{expt2}
\end{figure}

\begin{verbatim}
par(mfrow=c(1,2))
plot(X.hclust, main="X.hclust",sub="",xlab="")
plot(X.hclustPL.sq.wardD, main="X.hclustPL.sq.wardD",sub="",xlab="")
\end{verbatim}

\subsection{Experiment 3: Non-Ward Result Produced by {\tt hclust} and
{\tt hclust.PL}}

In this experiment, with different (non-squared valued) input, we achieve 
a well-defined hierarchical clustering, but one that differs from Ward.  
Code used is as follows, with output shown.  

\begin{verbatim}

# EXPERIMENT 3 ----------------------------------
X.hclustPL.wardD = hclust.PL(dist(y),method="ward.D")
X.hclust.nosq = hclust(dist(y),method="ward")

sort(X.hclustPL.wardD$height)
0.1573864 0.2422061 0.2664122 0.2901741 0.3030634 
0.3083869 0.3589344 0.3832023 0.4018957 0.5988721 
0.7443850 0.7915592 0.7985444 0.8016877 0.8414950 
0.9273739 1.4676446 2.2073106 2.5687307

sort(X.hclust.nosq$height)
0.1573864 0.2422061 0.2664122 0.2901741 0.3030634 
0.3083869 0.3589344 0.3832023 0.4018957 0.5988721 
0.7443850 0.7915592 0.7985444 0.8016877 0.8414950 
0.9273739 1.4676446 2.2073106 2.5687307

\end{verbatim}

This points to:
{\tt hclustPL.wardD} with {\tt method="wardD"} option
being the same as:
{\tt hclust} with {\tt method="ward"} option.

Note: there is no squaring of inputs in the latter, 
nor in the former either.  The clustering levels produced 
in this experiment using non-squared distances as input 
differ from, and are not monotonic relative to, those produced
in Experiments 1 and 2. 

Figure \ref{expt3} displays the outcome, and we see
the same visual result in both cases.   
This is the code used to produce Figure \ref{expt3}:

\begin{figure}
\begin{center}
\includegraphics[width=16cm]{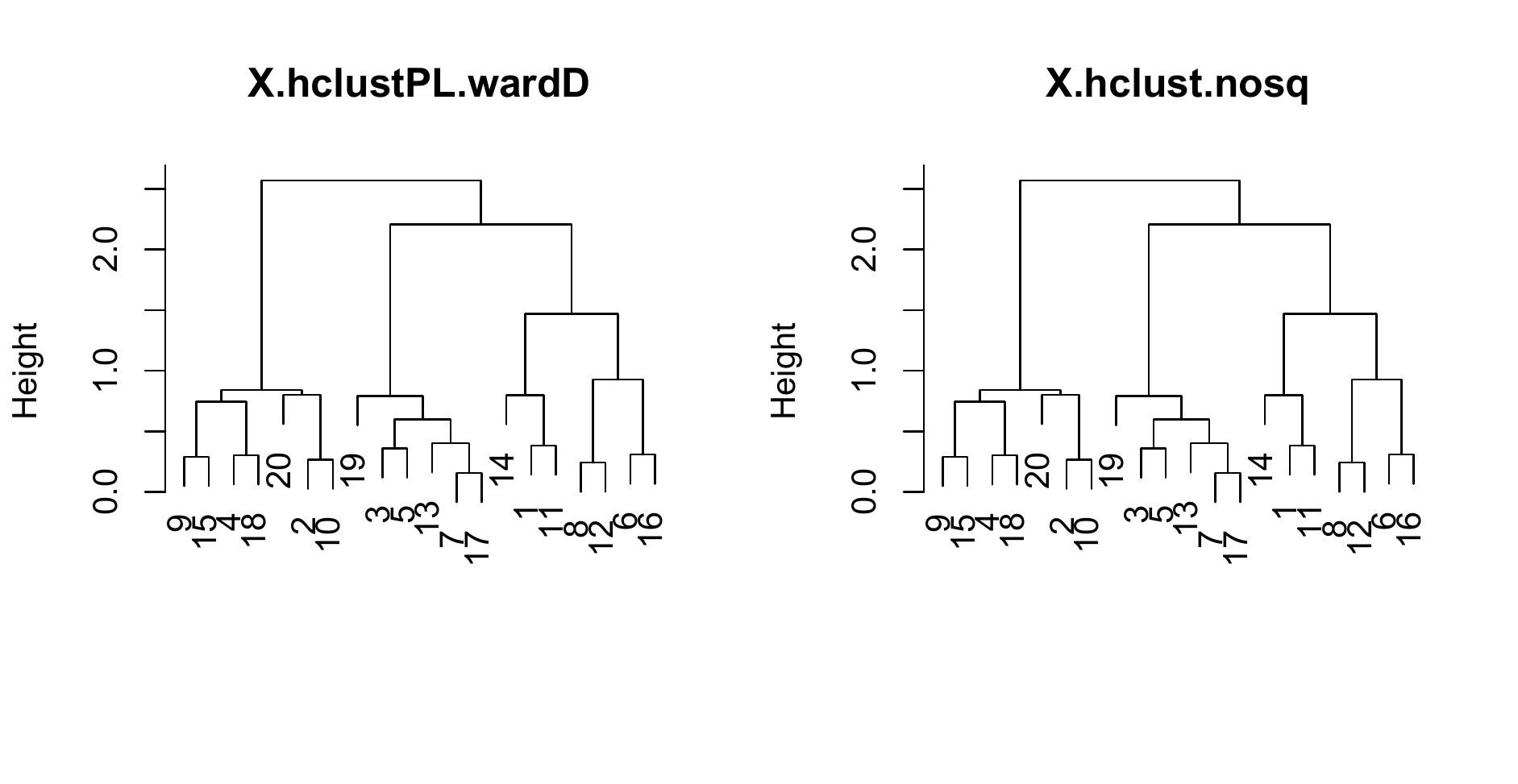}
\end{center}
\caption{Experiment 3 outcome.}
\label{expt3}
\end{figure}

\begin{verbatim}
par(mfrow=c(1,2))
plot(X.hclustPL.wardD, main="X.hclustPL.wardD",sub="",xlab="")
plot(X.hclust.nosq, main="X.hclust.nosq",sub="",xlab="")
\end{verbatim}

\subsection{Experiment 4: Modifying Inputs and Options so that Ward1 
Output is 
Identical to Ward2 Output}

In this experiment, given the formal equivalences of the Ward1 and 
Ward2 implementations in  sections \ref{sect31} and \ref{sect32}, 
we show how to bring about identical output.  We do this by squaring 
or not squaring input dissimilarities, and by playing on the options 
used.  

\begin{verbatim}
> # EXPERIMENT 4 ----------------------------------
X.hclust = hclust(dist(y)^2, method="ward")
X.hierclustPL.sq.wardD = hclust.PL(dist(y)^2, method="ward.D")
X.hclustPL.wardD2 = hclust.PL(dist(y), method="ward.D2")
X.agnes.wardD2 = agnes(dist(y),method="ward")

\end{verbatim}

We will ensure that the node heights in the tree are in 
``distance'' terms, i.e.\ in terms of the initial, unsquared
Euclidean distances as used in this article.  Of course, 
the agglomerations redefine such distances to be dissimilarities.
Thus it is with unsquared dissimilarities that we are concerned.  

While these dissimilarities are inter-cluster measures,
defined in any given partition, on the other hand the
inter-node measures that are defined on the tree are 
ultrametric. 

%
%
%

\begin{verbatim}
sort(sqrt(X.hclust$height))
0.1573864 0.2422061 0.2664122 0.2901741 0.3030634 
0.3083869 0.3589344 0.3830281 0.3832023 0.5753823 
0.6840459 0.7258152 0.7469914 0.7647439 0.8042245 
0.8751259 1.2043397 1.5665054 1.8584163

sort(sqrt(X.hierclustPL.sq.wardD$height))
0.1573864 0.2422061 0.2664122 0.2901741 0.3030634 
0.3083869 0.3589344 0.3830281 0.3832023 0.5753823
0.6840459 0.7258152 0.7469914 0.7647439 0.8042245 
0.8751259 1.2043397 1.5665054 1.8584163

sort(X.hclustPL.wardD2$height)
0.1573864 0.2422061 0.2664122 0.2901741 0.3030634 
0.3083869 0.3589344 0.3830281 0.3832023 0.5753823
0.6840459 0.7258152 0.7469914 0.7647439 0.8042245 
0.8751259 1.2043397 1.5665054 1.8584163

sort(X.agnes.wardD2$height)
0.1573864 0.2422061 0.2664122 0.2901741 0.3030634 
0.3083869 0.3589344 0.3830281 0.3832023 0.5753823
0.6840459 0.7258152 0.7469914 0.7647439 0.8042245 
0.8751259 1.2043397 1.5665054 1.8584163
 
\end{verbatim}

There is no difference of course between sorting 
the squared agglomeration or height levels, versus sorting 
them and then squaring them.  Consider the following 
examples, the first repeated from the foregoing (Experiment 4)
batch of results. 

\begin{verbatim}
sqrt(sort(X.hierclustPL.sq.wardD$height))
0.1573864 0.2422061 0.2664122 0.2901741 0.3030634 
0.3083869 0.3589344 0.3830281 0.3832023 0.5753823
0.6840459 0.7258152 0.7469914 0.7647439 0.8042245 
0.8751259 1.2043397 1.5665054 1.8584163
 
sqrt(sort(X.hierclustPL.sq.wardD$height))
0.1573864 0.2422061 0.2664122 0.2901741 0.3030634 
0.3083869 0.3589344 0.3830281 0.3832023 0.5753823
0.6840459 0.7258152 0.7469914 0.7647439 0.8042245 
0.8751259 1.2043397 1.5665054 1.8584163
 
\end{verbatim}


Our Experiment 4 points to:

\begin{description}
\item[ ] output of {\tt hclustPL.wardD2}, with the {\tt 
method="ward.D2"} option
\item[ ] or
\item[ ] output of {\tt agnes}, with the {\tt method="ward"}
option
\item[ ] being the same as both of the following with node
heights square rooted:
\item[ ] {\tt hclust}, with the {\tt "ward"} option on squared input,
\item[ ] 
{\tt hclust.PL}, with the {\tt method="ward.D"} option on squared input.
\end{description}

Figure \ref{expt4} displays two of these outcomes, and we see
the same visual result in both cases, in line with the numerical
node (or agglomeration) ``height'' values.   
This is the code used to produce Figure \ref{expt4}:

\begin{figure}
\begin{center}
\includegraphics[width=16cm]{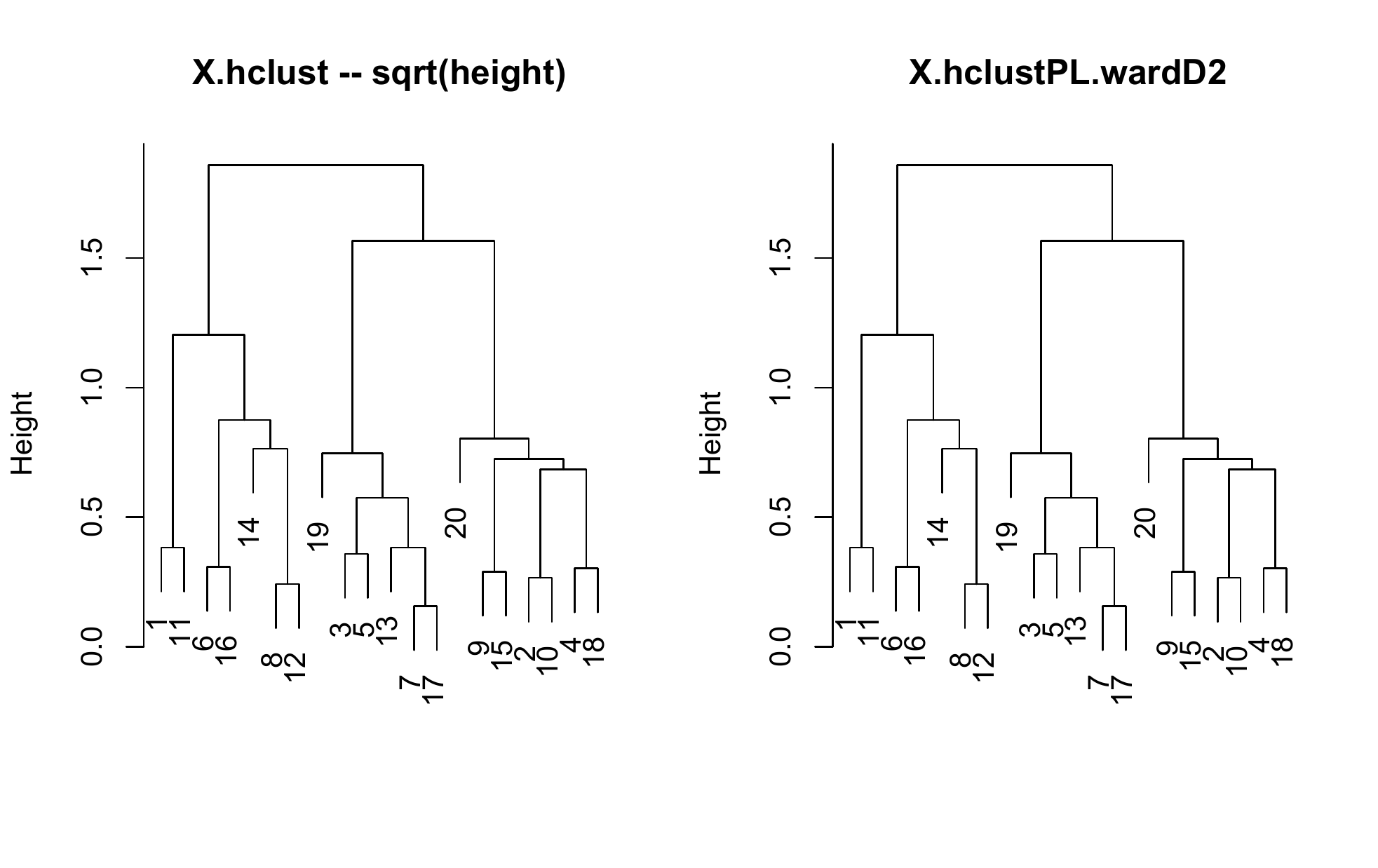}
\end{center}
\caption{Experiment 4 outcome.}
\label{expt4}
\end{figure}



\begin{verbatim}
par(mfrow=c(1,2))
temp <- X.hclust
temp$height <- sqrt(X.hclust$height)
plot(temp, main="X.hclust -- sqrt(height)", sub="", xlab="")
plot(X.hclustPL.wardD2, main="X.hclustPL.wardD2", sub="", xlab="")
\end{verbatim}

\section{Discussion}

\subsection{Where Ward1 and Ward2 Implementations Lead to an Identical Result}

A short discussion follows on the implications of this work. 
In Experiments 1 and 2, we see the crucial importance of inputs (squared or 
not) and options used.  
We set out, with Experiment 1, to implement Ward2.  With Experiment 2, we set 
out to implement Ward1.  Experiment 3 shows how easy it is to modify either 
implementation, Ward1 or Ward2, to get another (well defined) 
non-Ward hierarchy.

Finally Experiment 4 shows the underlying equivalence of the Experiment 1 
and Experiment 2 results, i.e.\ respectively the Ward2 and Ward1 
implementations.  

On looking closely at the Experiment 1 and Experiment 2 figures, Figures 
\ref{expt1} and \ref{expt2}, we can see that the morphology of the dendrograms
is the same.  However the cluster criterion values -- the node heights --
are not the same.   

From section \ref{sect32}, Ward2 implementation, 
the cluster criterion value is most naturally 
the square root of the same cluster criterion value as used in section
\ref{sect31}, Ward1 implementation.  From a dendrogram morphology viewpoint,
this is not important because one morphology is the same as the other
(as we have observed above). 
From an optimization viewpoint (section \ref{optim}), it plays no role
either since one optimant is monotonically related to the other.  

\subsection{How and Why the Ward1 and Ward2 Implementations Can Differ}

Those were reasons as to why it makes no difference to choose  
the Ward1 implementation versus the Ward2 implementation.   Next,
we will look at some practical differences.  

Looking closer at forms of the criterion in (\ref{wardcrit2}) and 
(\ref{wardcrit1}) in section \ref{optim} -- and contrasting these
forms of the criterion with the input dissimilarities in sections 
\ref{sect31} (Ward1) and \ref{sect32} (Ward2) leads us to the 
following observation.  The Ward2 criterion values are ``on a scale 
of distances'' whereas the Ward1 criterion values are ``on a scale
of distances squared''.   Hence to make direct comparisons between 
the ultrametric distances read off a dendrogram, and compare 
them to the input distances, it is preferable to use the Ward2 form 
of the criterion.   Thus, the use of cophenetic correlations can 
be more directly related to the dendrogram produced.   
Alternatively, with the Ward1 form of the criterion, we can 
just take the square root of the dendrogram node heights.  This 
we have seen in the generation of Figure \ref{expt4}. 


\subsection{Other Implementations Based on the Lance-Williams Update 
Formula}

The above algorithm can be used for  ``stored dissimilarity'' and 
``stored data'' implementations, a distinction first made in Anderberg
(1973).   The latter is where the dissimilarity matrix is not 
used, but instead the dissimilarities are created on the fly.  

Murtagh (2000) has implementations of the latter, ``stored data'', 
implementations (programs hc.f, HCL.java, see Murtagh, 2000) 
as well as the former, ``stored data'' (hcon2.f).  For both, Murtagh (1985)
lists the formulas.   The nearest neighbor and reciprocal nearest 
neighbor algorithms can be applied to bypass the need for a strict sequencing
of the agglomerations.   See Murtagh (1983, 1985).   These algorithms
provide for provably worst case $O(n^2)$ implementations, as first
introduced in 
de Rham and Juan, and published in, respectively, 1980 and 1982.  Cluster 
criteria such as Ward's method must respect Bruynooghe's reducibility 
property if they are to be inversion-free (or with monotonic variation 
in cluster criterion value through the sequence of agglomerations).  
Apart from computational reasons, the other major advantage
of such algorithms (nearest neighbor chain, reciprocal nearest neighbor)
is use in distributed computing (including virtual memory)
environments. 

\section{Conclusions}

Having different very close implementations that differ by just a 
few lines of code (in any high level language), yet claiming to 
implement a given method, is confusing for the learner, for the 
practitioner and even for the specialist.  
In this work, we have first of all reviewed all relevant background. 
Then we have laid out in very clear terms the two, differing 
implementations.  Additionally, with differing inputs, and with 
somewhat different processing driven by options set by the user,
in fact our two different implementations had the appearance of
being quite different methods.  

Two algorithms, Ward1 and Ward2, are found in the literature and software, both announcing that they implement the Ward (1963) clustering method. When applied to the same distance matrix D, they produce different results. This article has shown that when they are applied to the same dissimilarity matrix D, only Ward2 minimizes the Ward clustering criterion and produces the Ward method. The Ward1 and Ward2 algorithms can be made to optimize the same criterion and produce the same clustering topology by using Ward1 with D-squared and Ward2 with D. Furthermore, taking the square root of the clustering levels produced by Ward1 used with D-squared produces the same clustering levels as Ward2 used with D.
The constrained clustering package of Legendre (2011), \verb+const.clust+,
derived from \verb+hclust+ in R, offers both the Ward1 and
Ward2 options.  

We have shown in this article how close these two implementations are, 
in fact.  Furthermore we discussed in detail what the implications 
are for the few, differing lines of code.  
Software developers who only offer the Ward1 algorithm are encouraged to explain 
clearly how the Ward2 output is to be obtained, as described in the previous paragraph.

\section*{References}

\begin{enumerate}

\item Anderberg, M.R., 1973. Cluster Analysis for Applications, Academic. 
\item Batagelj, V., 1988.  
Generalized Ward and related clustering problems, in 
H.H. Bock, ed., Classification and Related Methods of Data Analysis, 
North-Holland, pp. 67--74. 

\item Benz\'ecri, J.P., 1976. L'Analyse des Donn\'ees, Tome 1, La Taxinomie, 
Dunod, (2nd edn.; 1973, 1st edn.). 


\item Cailliez, F. and Pag\`es, J.-P., 1976. Introduction \`a l'Analyse des 
Donn\'ees, SMASH (Soci\'et\'e de Math\'ematiques Appliqu\'ees et Sciences
Humaines).  

\item Fisher, R.A., 1936. The use of multiple measurements in taxonomic 
problems.  Annals of Eugenics, 7, 179--188.  

\item Jain, A.K. and Dubes, R.C., 1988. Algorithms for Clustering Data, 
Prentice-Hall.

\item Jambu, M., 1978.  
Classification Automatique pour l'Analyse des Donn\'ees.
I. M\'ethodes et Algorithmes, Dunod.

\item Jambu, M., 1989. Exploration Informatique et Statistique des Donn\'ees, 
Dunod.

\item Kaufman, L. and Rousseeuw, P.J., 1990. 
Finding Groups in Data: An Introduction
to Cluster Analysis, Wiley.  

\item Lance, G.N. and Williams, W.T., 1967. A general theory of classificatory 
sorting strategies.  1. Hierarchical systems, The Computer Journal, 9, 
4, 373--380. 

\item Legendre, P. and Fortin, M.-J., 2010. Comparison of the Mantel test 
and alternative approaches for detecting complex relationships in 
the spatial analysis of genetic data.  Molecular Ecology Resources,
10, 831--844.  

\item Legendre, P., 2011.  \verb+const.clust+, 
R package to compute space-constrained or 
time-constrained agglomerative clustering. \\
http://www.bio.umontreal.ca/legendre/indexEn.html

\item Legendre, P. and Legendre, L., 2012. Numerical Ecology, 3rd ed., 
Elsevier.

\item Le Roux, B. and Rouanet, H., 2004. Geometric Data Analysis: From 
Correspondence Analysis to Structured Data Analysis, Kluwer. 

\item Murtagh, F., 1983. A survey of recent advances in hierarchical 
clustering algorithms, The Computer Journal, 26, 354--359.

\item Murtagh, F., 1985. 
Multidimensional Clustering Algorithms, Physica-Verlag.

\item Murtagh, F., 2000. Multivariate data analysis software and resources, \\
http://www.classification-society.org/csna/mda-sw

\item Sz\'ekely, G.J. and Rizzo, M.L., 2005. 
Hierarchical clustering via joint between-within distances: extending 
Ward's minimum variance method, Journal of Classification, 22 (2), 
151--183.
\item Ward, J.H., 1963. 
Hierarchical grouping to optimize an objective function,
Journal of the American Statistical Association, 58, 236--244.
\item Wishart, D., 1969. An algorithm for hierachical classifications, 
Biometrics 25, 165--170.
\item XploRe, 2007. 
http://fedc.wiwi.hu-berlin.de/xplore/tutorials/xaghtmlnode53.html
\end{enumerate}

\end{document}